\pdfoutput=1

\documentclass[11pt]{article}

\usepackage[preprint]{acl}

\usepackage{times}
\usepackage{latexsym}

\usepackage[T1]{fontenc}

\usepackage[utf8]{inputenc}

\usepackage{microtype}

\usepackage{inconsolata}

\usepackage{graphicx}

\usepackage{graphicx}
\usepackage{amsmath}

\usepackage{tabularx}

\usepackage{booktabs} 
\usepackage{array}    
\usepackage{multirow}  
\usepackage{float}

\usepackage{adjustbox}

\title{KGPA: Robustness Evaluation for Large Language Models via Cross-Domain Knowledge Graphs}

\author{Aihua Pei \\Waseda University\\
  aika@fuji.waseda.jp\And
    Zehua Yang \\ Waseda University \\ yangzehua@akane.waseda.jp\And
    Shunan Zhu \\ Waseda University \\ shunan-zhu@ruri.waseda.jp
    \AND
    Ruoxi Cheng \\ Southeast University \\ 213200761@seu.edu.cn \And
    Ju Jia \\ Southeast University \\ jiaju@seu.edu.cn \And
    Lina Wang \\ Wuhan University \\lnwang@whu.edu.cn}

\begin{document}
\maketitle
\begin{abstract}
Existing frameworks for assessing robustness of large language models (LLMs) overly depend on specific benchmarks, increasing costs and failing to evaluate performance of LLMs in professional domains due to dataset limitations. This paper proposes a framework that systematically evaluates the robustness of LLMs under adversarial attack scenarios by leveraging knowledge graphs (KGs). Our framework generates original prompts from the triplets of knowledge graphs and creates adversarial prompts by poisoning, assessing the robustness of LLMs through the results of these adversarial attacks. We systematically evaluate the effectiveness of this framework and its modules. Experiments show that adversarial robustness of the ChatGPT family ranks as GPT-4-turbo > GPT-4o > GPT-3.5-turbo, and the robustness of large language models is influenced by the professional domains in which they operate.
\end{abstract}

\section{Introduction}

Large language models (LLMs) have garnered significant attention due to their exceptional performance across various natural language processing (NLP) tasks. However, as these models are widely applied in critical domains, they also face the risk of adversarial attacks triggered by prompts. Adversarial attacks aim to mislead models into making incorrect judgments through carefully designed prompts, potentially causing severe damage to users. Therefore, it is necessary to assess the robustness of models against adversarial attacks using adversarial robustness evaluations.

Existing adversarial robustness evaluation frameworks for large language models (LLMs), like AdvGLUE and PromptAttack, use specialized benchmark datasets that require extensive manual annotation \cite{wang2021adversarial, lin2021truthfulqa}. This not only limits their applicability but also increases operational costs. Moreover, when LLMs are used in specialized domains such as medicine or biology, the mismatch between generic benchmark datasets and the specific context can lead to inaccurate robustness evaluations. These limitations decrease practicality of the frameworks and complicate the robustness evaluation of LLMs.

\begin{figure}[ht]
    \centering
    \includegraphics[width=1 \linewidth]{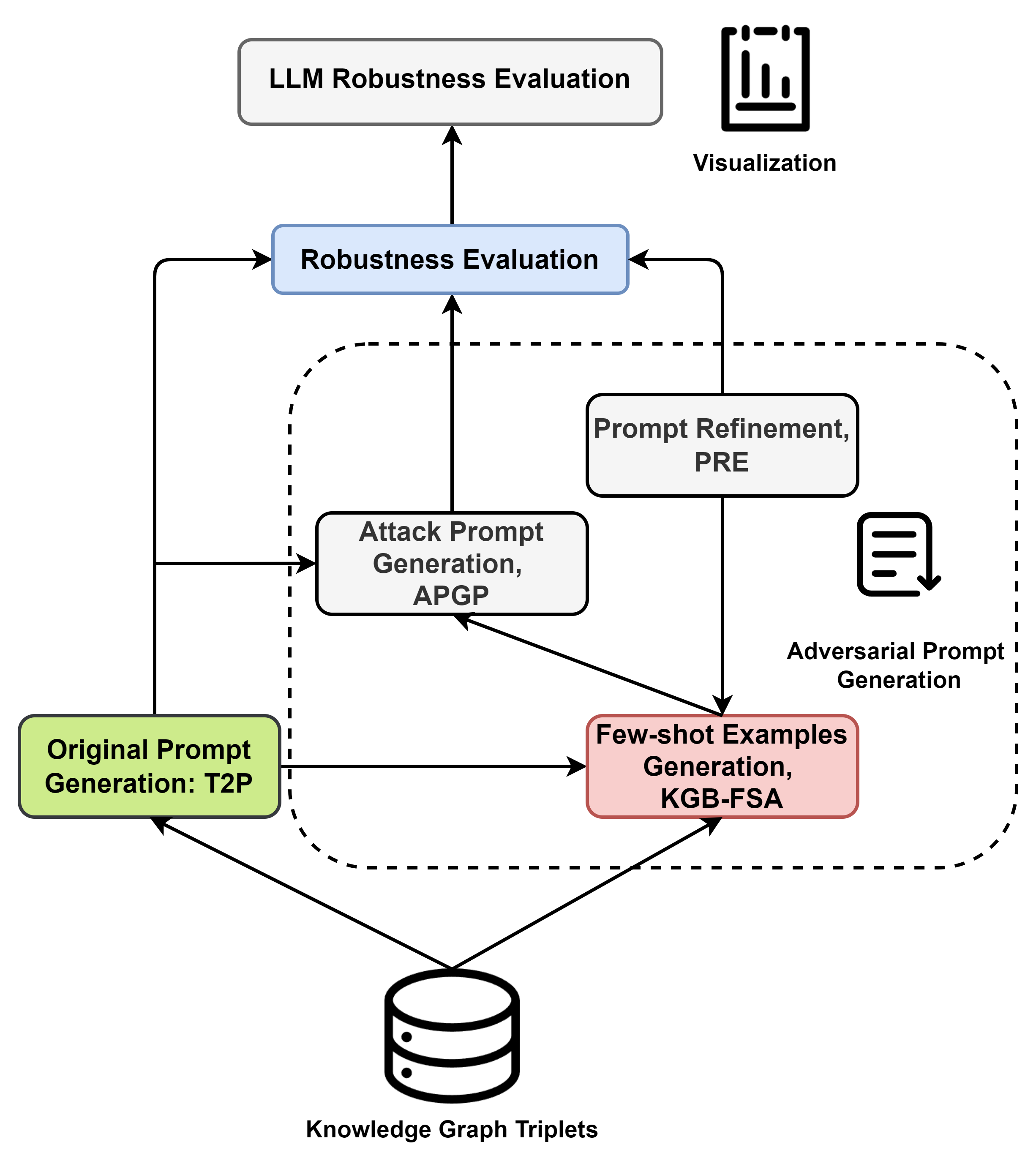}
    \caption{Framework of Knowledge Graph Based PromptAttack (KGPA)}
    \label{fig:enter-label}
\end{figure}

This paper proposes an adversarial attack framework (Knowledge Graph Based PromptAttack, KGPA) utilizing knowledge graphs to generate and poison prompts from graph triplets, evaluating LLM robustness. The generation of adversarial prompts is meticulously crafted to optimize quality and evaluation effectiveness. We apply this framework to generate prompts from both general and specialized domain knowledge graphs, evaluating the resilience of multiple LLMs under adversarial attack conditions. Specifically, our contributions include:

\begin{itemize}
\item We propose a new framework that efficiently generates original and adversarial prompts from triplets in knowledge graphs without relying on specially constructed benchmark datasets. This framework uses these prompts in adversarial attacks to evaluate the robustness of LLMs.

\item For the KGPA framework, we meticulously design modules for generating original prompts, producing and optimizing adversarial prompts, and selecting the most effective adversarial prompts. We evaluate these modules through experiments and explore how various settings within the modules affect the results of adversarial attacks and robustness evaluations.

\item We confirm that the robustness of large language models is influenced by the scope of knowledge corresponding to the prompts. Although the tested large language models have roughly the same robustness ranking under different knowledge graphs, the robustness of the same large language model measured on general or specialized domain knowledge graphs is not similar.
\end{itemize}

\section{Related Works}
\subsection{Robustness Evaluation of LLMs}
Large language models (LLMs), such as the ChatGPT family and the Llama family, have attracted much attention for their excellent performance in a variety of natural language processing tasks \cite{touvron2023llama,brown2020language}. However, as these models are widely used in critical domains applications, assessing their robustness has also become a hot research topic. There are four main streams of work \cite{li2023survey, ailem2024examining, zhuo2023robustness} on robustness research: robustness under distribution shift \cite{yang2023glue}, robustness to adversarial attacks \cite{wang2023robustness, zhu2023promptbench}, robustness to prompt formats and instruction templates \cite{mizrahi2023state,voronov2024mind,weber2023mind} and robustness to
dataset bias \cite{gururangan2018annotation, niven2019probing, le2020adversarial}. Our work focus on evaluating robustness to adversarial attacks of LLMs.

Adversarial attacks aim to mislead the model to make wrong judgments through well-designed inputs, while adversarial robustness evaluation attempts to determine and enhance robustness of the model to these attacks. Current robustness evaluation frameworks for LLMs are mainly based on specially constructed benchmark datasets (e.g. the GLUE dataset \cite{wang2018glue} and ANLI dataset \cite{nie2020adversarial}) for evaluating natural language comprehension capabilities of LLMs \cite{goel2021robustness}.

 AdvGLUE and AdvGLUE++ \cite{wang2023decodingtrust} are two frameworks specifically designed to evaluate the adversarial robustness of language models. These frameworks challenge the ability to make judgments under complex and subtle semantic changes by providing a series of adversarial samples of models. AdvGLUE++ is a further extension of AdvGLUE that introduces more adversarial samples, especially for new emerging LLMs such as the Alpaca and Vicuna families \cite{taori2023alpaca, vicuna2023}.PromptAttack enhance the attack power by ensembling adversarial examples at different perturbation levels \cite{xu2023llm}. These evaluation frameworks exhibit a common feature: testing and improving the robustness of the model by constructing inputs that may cause the model to misjudge. These inputs include both subtle textual modifications and complex semantic transformations, aiming to comprehensively evaluate robustness of the model to various challenges that may be encountered in real-world applications. 

\subsection{Attack Prompt Generation from KG}

In evaluating robustness of LLMs, we need to know whether they have such knowledge and whether they can accurately express their knowledge. Knowledge graph can help us generate attack prompt with different diversities and complexities. Knowledge graph (KG) is a graph structure for representing knowledge, where nodes represent entities or concepts and edges represent relationships between these entities or concepts.

Some works use different methods to utilize triplet from Knowledge Graphs generating questions \cite{seyler2017knowledge, kumar2019difficulty, chen2023toward}. Some works utilize the ability of LLMs to generate
questions from KGs \cite{guo2022dsm,axelsson2023using}. Recent works \cite{luo2023systematic, luo2024biaskg} also discussed on evaluating factual knowledge of LLMs with the diverse and well-coverage questions generated from KGs and how KGs can be used to induce bias in LLMs.

\subsection{Few-Shot Attack Strategy}

As the application of machine learning models, especially large language models, in various tasks becomes popular, the few-shot attack strategy has also received considerable attention \cite{logan2021cutting, meng2024advancing}. It utilizes adversarial samples to evaluate and improve ability of the models to resist attacks. Adversary samples are malicious samples that can cause models to make incorrect predictions, where an adversary adds a small perturbation to the original benign sample that is difficult for humans to detect, thus creating an adversary sample that can deceive the target model.

The few-shot attack strategy is designed to efficiently execute attacks using only a small number of samples, which is particularly important for evaluating newly emerging LLMs, typically require large amounts of data for training and testing. At the heart of the strategy is the fact that precisely designed attacks can reveal potential weaknesses in a model even with a limited number of samples. 


\section{Methodology}


This section presents our Knowledge Graph based PromptAttack (KGPA) framework, which converts knowledge graph triplets into original prompts, modifies them into adversarial prompts, selects suitable ones for attacks, and evaluates the robustness of large language models using specific metrics.

\subsection{Original Prompt Generation}

The task of generating original prompts is accomplished by the Triplets to Prompts (T2P) module. Our framework leverages facts stored in a knowledge graph, organized as triplets (Subject, Predicate, Object). It automatically generates original prompts with three labels: "true" for prompts representing correct knowledge, "entity\textunderscore{}error" for prompts with incorrect Subjects or Objects, and "predicate\textunderscore{}error" for prompts with incorrect Predicates.

The T2P module begins by altering certain triplets, swapping the Subject, Object, or Predicate to create new triplets with either entity errors (errors in the Subject or Object) or predicate errors. As a result, the sentences generated by the T2P module are tagged with one of three labels: "true", "entity\textunderscore{}error" or "predicate\textunderscore{}error".

For converting triplets into prompts, we offer two strategies: template-based and LLM-based. This section details these strategies:\\
\textbf{Template-Based Strategy} Many knowledge graphs use templates to illustrate connections between subjects and objects in sentences. For example, the template for "place of birth" is:

\begin{center}
\textit{[X]'s place of birth is [Y].}
\end{center}

\noindent{}Here, [X] and [Y] represent the subject and object of the triplet. Replacing these placeholders with specific names, like [X] as Isaac Newton and [Y] as England, the prompt becomes \textit{"Isaac Newton's place of birth is England."}\\
\textbf{LLM-based transformation strategie} The LLM-based transformation strategy involves feeding triplets from a knowledge graph (Subject, Predicate, Object) to a large language model, which then generates descriptive sentences based on these elements.

\begin{figure}[h]
    \centering
    \includegraphics[width=1 \linewidth]{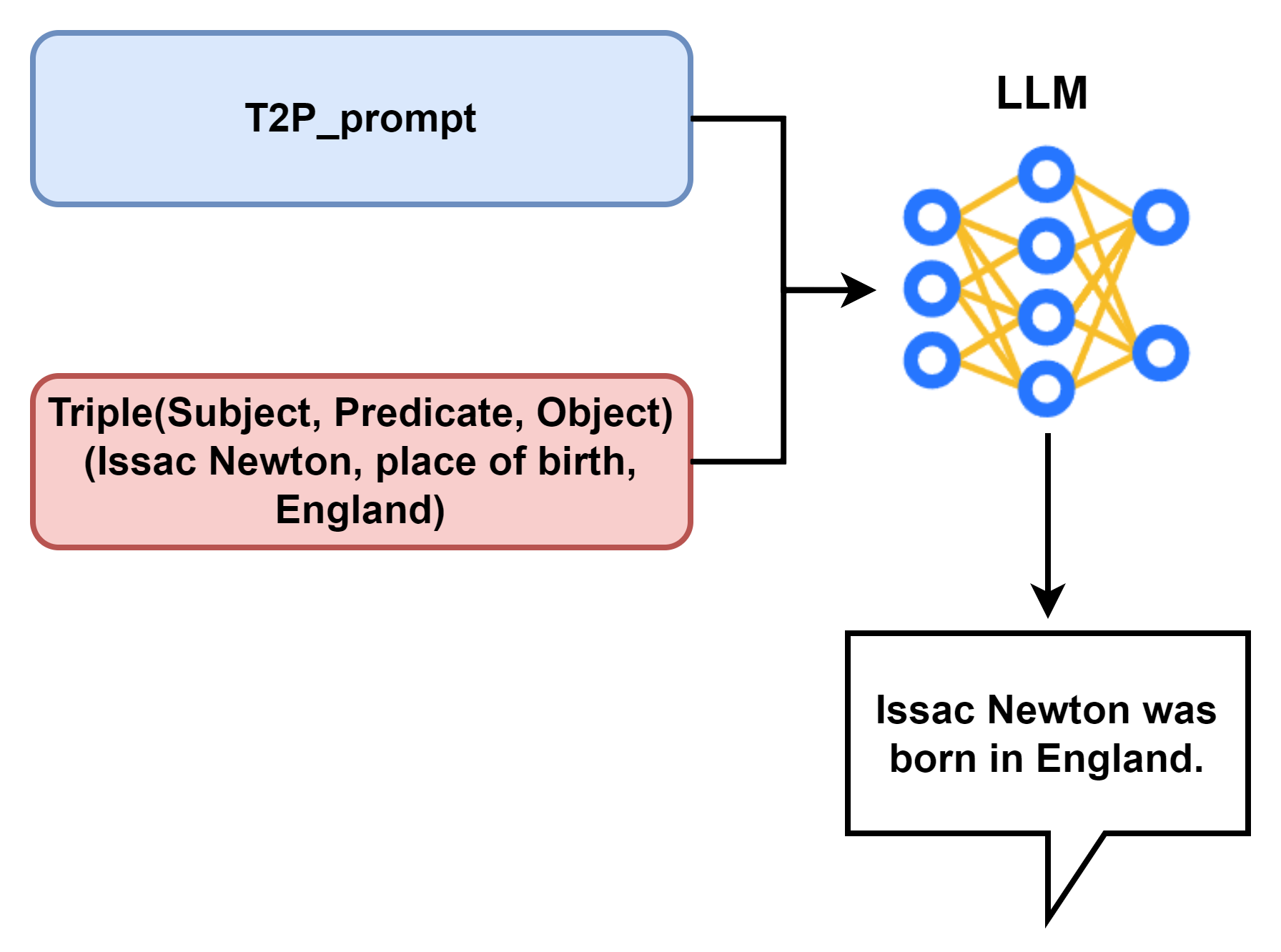}
    \caption{LLM-based transformation strategie example}
    \label{fig:t2p}
\end{figure}

\noindent{}The advantage of this transformation strategy is that it generates sentences of higher quality that are more easily understood by large language models.

\subsection{Adversarial Prompt Generation}
\textbf{KGB-FSA module.} This module generates a limited number of example prompts to implement a few-shot attack strategy within the APGP module. These example prompts must satisfy the following two conditions: their semantics should be sufficiently similar to the original prompts, and their classification by the large language model should differ from that of the original prompts, thus enabling effective adversarial attacks on the large language model. The KGB-FSA module comprehensively performs the process from converting triplets into sentences to using adversarial prompts to attack the large language model. It also records the prompts that successfully attack the large language model as example prompts. The basic architecture of the KGB-FSA module is illustrated in the following Figure 3.

\begin{figure}[h]
    \centering
    \includegraphics[width=1 \linewidth]{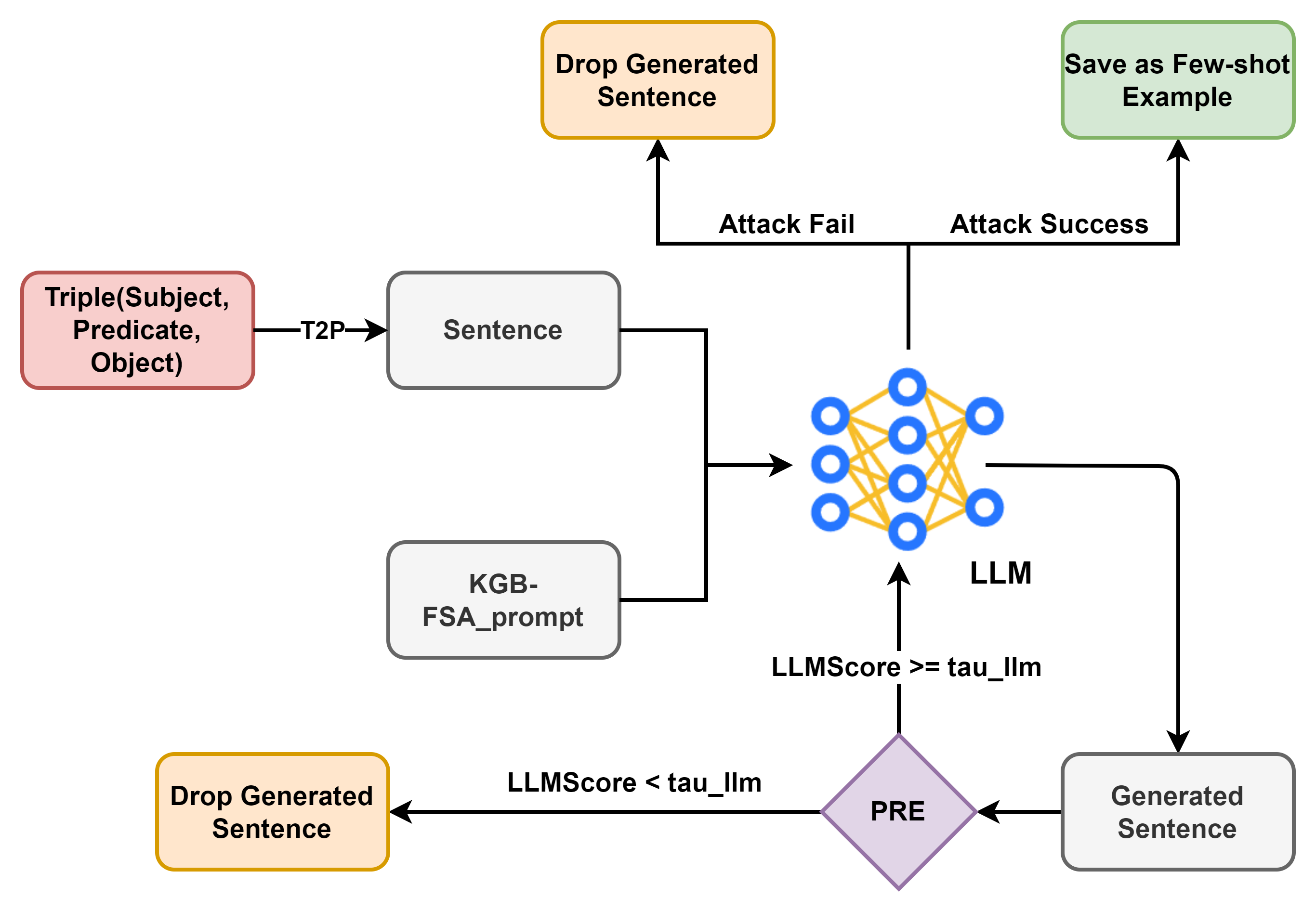}
    \caption{Knowledge graph-based few-shot attack strategy module (KGB-FSA)}
    \label{fig:enter-label}
\end{figure}

The design concept of this module is as follows: The sentences generated by the T2P module are first combined with the KGB-FSA\_prompt template and input into the large language model. The KGB-FSA\_prompt instructs the model to modify the original sentences so that the new sentences retain semantic consistency with the originals while resulting in different classification outcomes by the large language model. After the large language model generates the sentences, the Prompt Refinement Engine (PRE) filters out those that are not semantically consistent with the original sentences. The sentences that pass through the PRE module are then used to attack the large language model. Those that successfully attack the model are recorded as example prompts alongside the original sentences.

\noindent\textbf{Adversarial Prompt Generation Prompt (APGP).} This module randomly selects some of the original prompts generated by the T2P module and uses an APGP\_prompt, similar to the KGB-FSA\_prompt, to guide the large language model in modifying these original prompts. This process generates adversarial prompts that can be used to attack the model itself. Unlike the KGB-FSA\_prompt, the APGP\_prompt includes an optional few-Shot Attack Strategy, which allows for the inclusion of example prompts within the APGP\_prompt to provide reference points for the model in generating results. For the sentences generated by the model, the KGPA uses the PRE module to verify if they meet the required semantic similarity and quality standards. If they pass the PRE verification of module, the generated prompts are output as results; otherwise, the original prompts are output. Consequently, the final output of the APGP module consists of adversarial prompts. The basic structure of this module is illustrated in Figure 4.

\begin{figure*}[t]
    \centering
    \includegraphics[width= \textwidth]{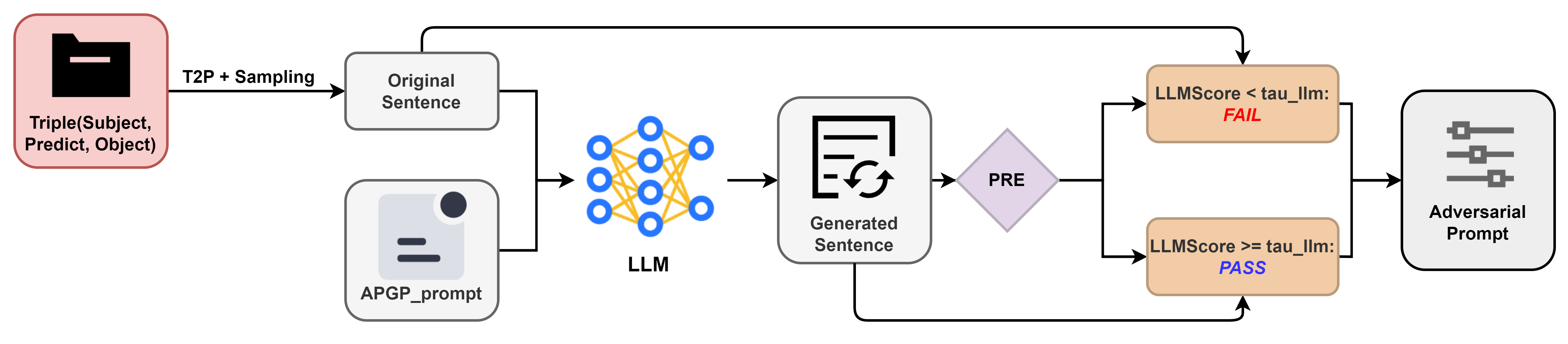}
    \caption{APGP: Adversarial prompt generation}
    \label{fig:enter-label}
\end{figure*}

\begin{figure}[h]
    \centering
    \includegraphics[width=1 \linewidth]{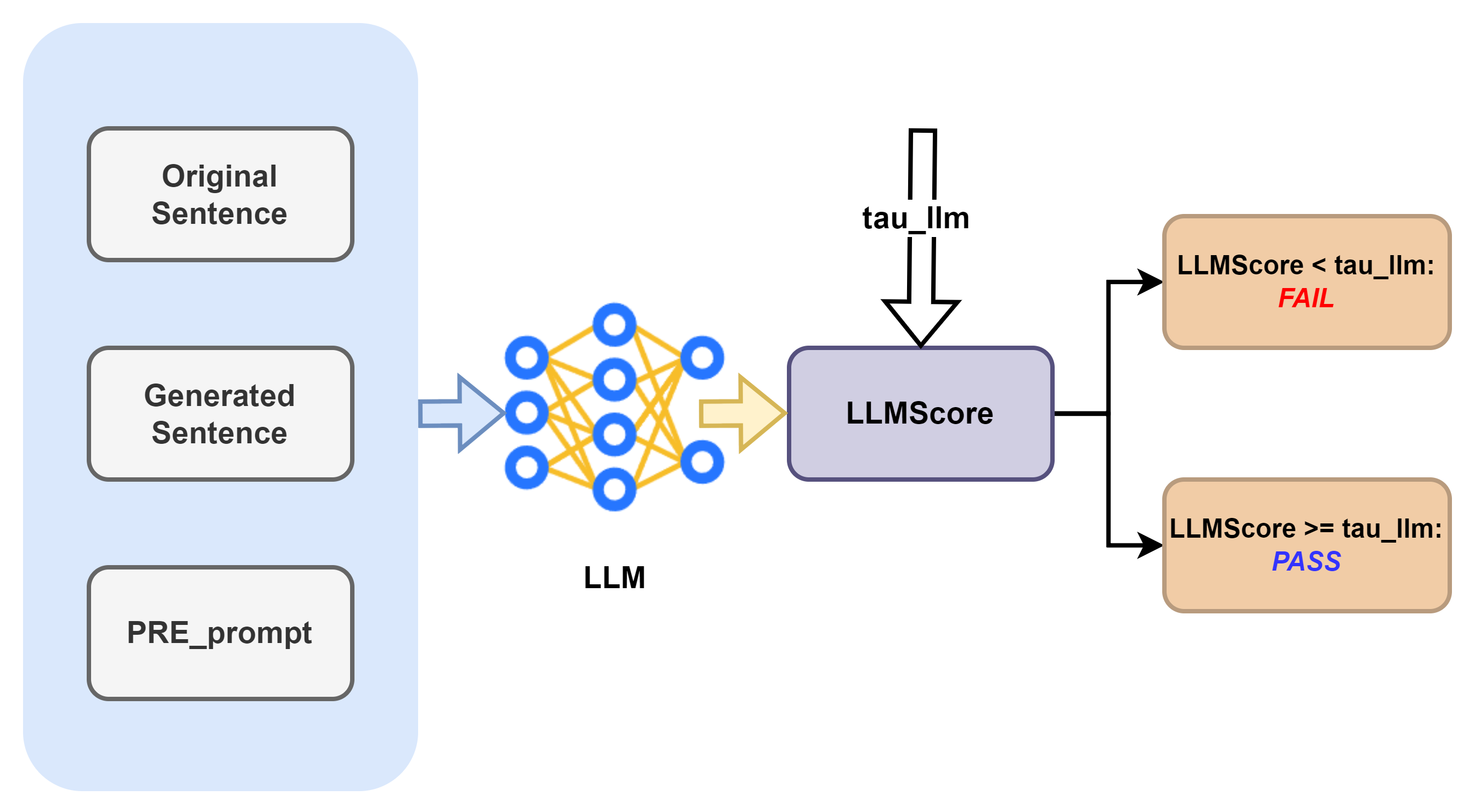}
    \caption{The basic architecture of prompt refinement engine (PRE) module}
    \label{fig:enter-label}
\end{figure}

\noindent\textbf{Prompt Refinement Engine (PRE).} We propose a new method LLMScore, offering several advantages over traditional metrics like Word Error Rate (WER) and BERTScore \cite{zhang2019bertscore}. WER measures the minimum number of substitutions, insertions, and deletions needed to transform one text into another, but it often fails to capture semantic similarities. For instance, it might assign a high error rate to sentences that are semantically close but lexically different. BERTScore improves by using cosine similarity of word embeddings to evaluate semantic similarity, but it can still misjudge sentences that differ significantly in context-specific meaning. LLMScore, however, leverages the large language model itself to assess the quality and semantic similarity of generated sentences compared to the original sentences. The score ranges from -1 to 1, with higher values indicating greater semantic similarity and quality. The PRE uses this score to filter generated sentences, comparing the LLMScore to a threshold, tau\_llm. If the LLMScore meets or exceeds this threshold, the sentence is considered semantically similar and of high quality. This approach ensures that the adversarial prompts generated are both effective and of high quality, enhancing the robustness evaluation of large language models. The basic structure of this module is illustrated in Figure 5.


\subsection{Robustness Evaluation Metrics}

After the APGP module generates adversarial prompts, we direct the large language model to classify both the original and the adversarial prompts. We assess model robustness using three key metrics: Natural Response Accuracy (NRA), Robust Response Accuracy (RRA), and Adversarial Success Rate (ASR). 
Assume the original set of prompts is $P$, and the adversarial set of prompts generated from $P$ is $P'$. Let $p$ be an element in $P$, with its corresponding element in $P'$ being $p'$. $L(p)$ represents the classification label given by the large language model $L$ to $p$, and $T(p)$ denotes the true label of $p$ and $p'$. $N(k)$ is the number of elements satisfying condition $k$. The formulas for the three key metrics of the prompt set $P$ on the large language model $L$ are as follows:

\noindent{}\textbf{Natural Response Accuracy (NRA)}. This method quantifies the ability of model to accurately classify original prompts, reflecting its understanding of the knowledge in the knowledge graph:
\begin{equation}
  \label{eq:NRA}
  NRA(P, L) = \frac{N(p \in P \& L(p) = T(p))}{N(p \in P)}
\end{equation}

\noindent{}\textbf{Robust Response Accuracy (RRA)}. This method evaluates the robustness of model by measuring its accuracy against adversarial prompts, indicating how well it can maintain classification integrity under potential disruptions:
\begin{equation}
  \label{eq:RRA}
  RRA(P, L) = \frac{N(p \in P \& L(p') = T(p))}{N(p \in P)}
\end{equation}

\noindent{}\textbf{Adversarial Success Rate (ASR)}. This method is a key metric for evaluating the robustness of LLMs when facing adversarial attacks. A high ASR indicates significant impact by the attacks, suggesting lower robustness, whereas a low ASR signifies a stronger resistance and higher robustness:
\begin{equation}
\label{eq:ASR}
\resizebox{.85\hsize}{!}{$
ASR(P, L) = \frac{N(p \in P \& L(p) = T(p) \& L(p') \neq T(p))}{N(p \in P \& L(p) = T(p))}
$}
\end{equation}

\section{Experiments}

\subsection{Arrangement}
\noindent{}The following two knowledge graphs are general domain knowledge graph datasets: Google-RE \cite{petroni-etal-2019-language} and T-REx \cite{elsahar-etal-2018-rex}. The following two are specialized domain knowledge graph datasets: UMLS \cite{Bodenreider2004TheUM} and WikiBio \cite{sung2021can}. The large language models of the ChatGPT family used for prompt adversarial attacks and robustness evaluation include GPT-3.5-turbo, GPT-4-turbo, and GPT-4o. In most experimental settings, the threshold for the Prompt Refinement Engine (PRE) module, tau\_llm, is set to 0.92.


\subsection{Robustness Evaluation of ChatGPT Family}
\textbf{Results and Analysis.} Figure 6 presents the average robustness test results for three large language models from the ChatGPT family across four knowledge graph datasets. The tests examine three distinct scenarios: (1) Performance of each model on each dataset using the template-based T2P module with a few-shot attack strategy. (2) Performance using the LLM-based T2P module with a few-shot attack strategy. (3) Performance using the template-based T2P module without the few-shot attack strategy. This comprehensive analysis highlights the differences in model robustness across various domains.


\noindent\textbf{ASR values.} 
In Figure 6, the ASR values indicate a robustness hierarchy of GPT models as follows: GPT-4-turbo > GPT-4o > GPT-3.5-turbo. This ranking corresponds to the robustness predictions made in this paper for these three large language models. OpenAI provides GPT-4o and GPT-3.5 as free services, highlighting their utility, whereas GPT-4 is a premium service, potentially reflecting its superior robustness.

GPT-4o, released after GPT-3.5, appears to trade some robustness for efficiency. This trade-off is evident in general domain datasets like T-REx and Google-RE, where GPT-4o significantly outperforms GPT-3.5-turbo in robustness against adversarial attacks. However, in specialized domains such as UMLS and WikiBio, GPT-4o and GPT-3.5-turbo show similar robustness, with GPT-4o even slightly underperforming in the WikiBio dataset. This suggests that while GPT-4o is well-suited for general applications, its effectiveness in specialized domains may be limited.

\begin{figure}[h]
    \centering
    \includegraphics[width=1\linewidth]{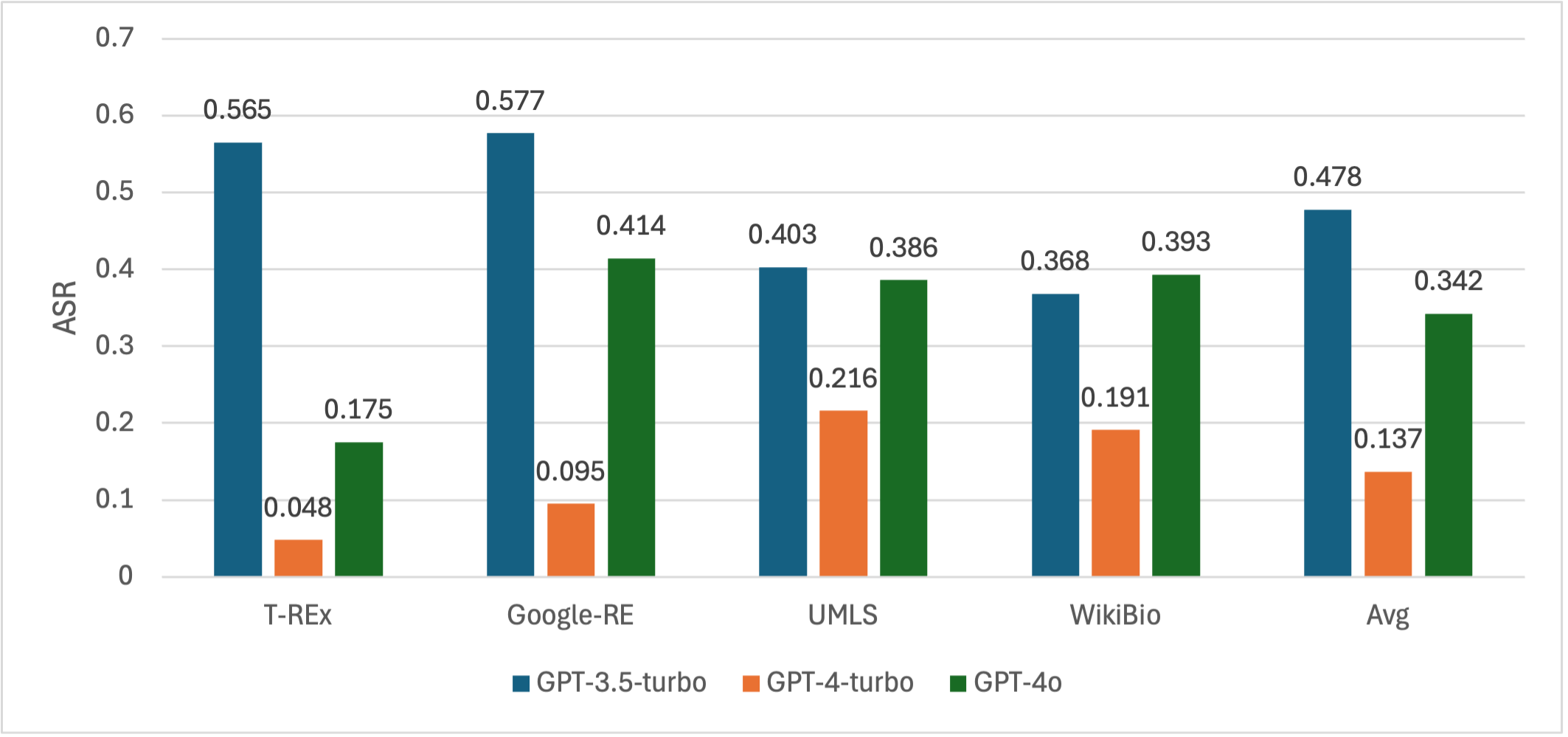}
    \caption{ASR values of the three LLMs on each knowledge graph}
    \label{fig:enter-label}
\end{figure}
The analysis also reveals consistent robustness within the same domain for each model, underscoring the models' domain-specific performance. GPT-3.5-turbo shows better robustness in specialized domains compared to general domains, while GPT-4-turbo and GPT-4o are more robust in general domains. This highlights the influence of model design and training data: GPT-4 family models are optimized for broader applicability, whereas GPT-3.5 may benefit more from specialized training.

\noindent\textbf{NRA values.} Figure 7 illustrates the NRA values of GPT-3.5-turbo and the GPT-4 family across four datasets, focusing on their performance under non-adversarial prompts. 
\begin{figure}[h]
        \centering
        \includegraphics[width=1\linewidth]{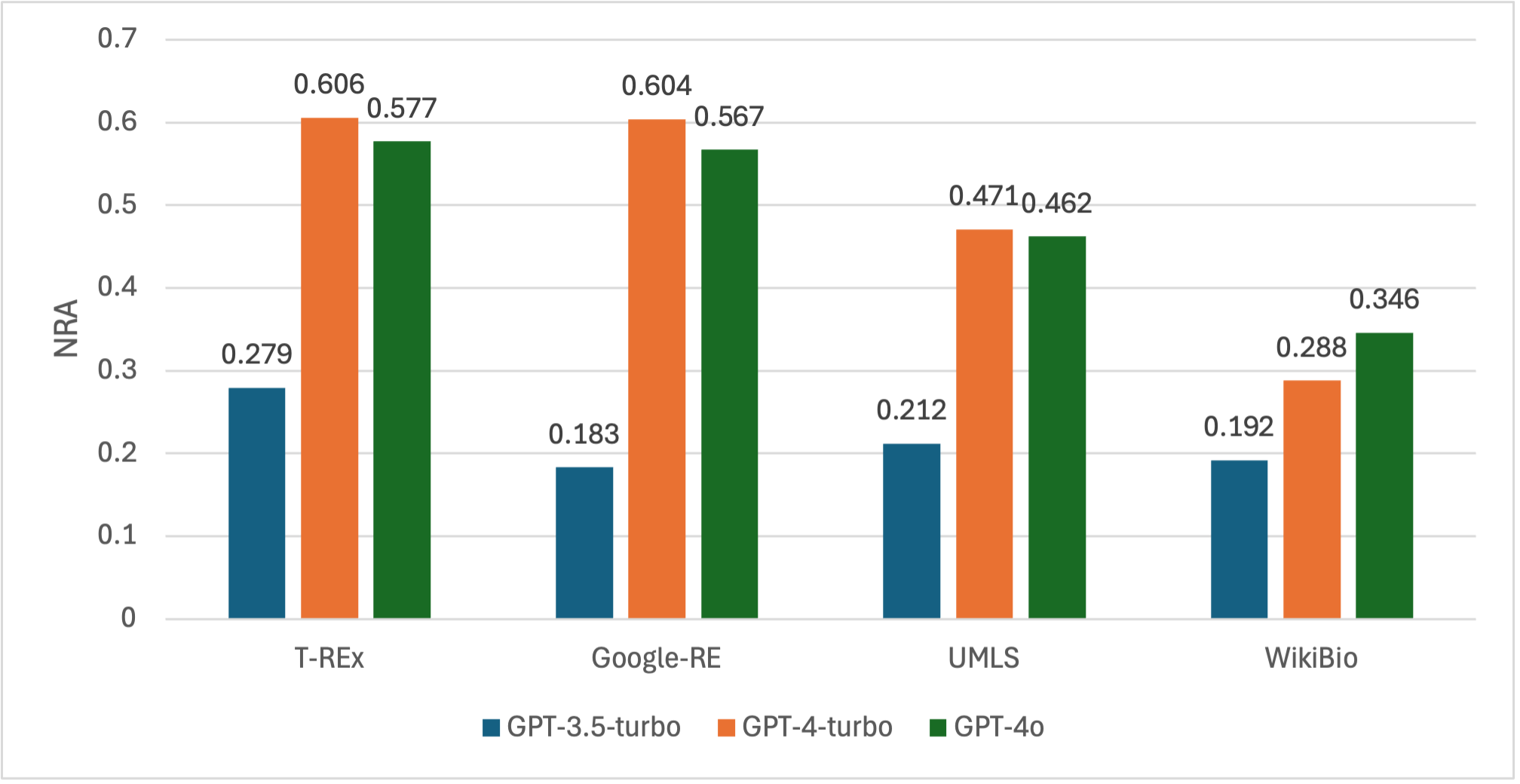}
        \caption{NRA values of the three LLMs on each knowledge graph}
        \label{fig:enter-label}
\end{figure}
The data reveals that all models generally perform better on general domain datasets compared to specialized domains, demonstrating a stronger command of broad knowledge. However, this trend is more pronounced in the GPT-4 family, highlighting its design focus on generalization. In contrast, GPT-3.5-turbo, which includes training on specialized fields, shows a closer performance gap between general and specialized knowledge domains. This distinction may underscore the influence of model design and training dataset selection on the models' capabilities in handling knowledge from different domains.

\noindent\textbf{RRA values.} In analyzing the performance of GPT-4-turbo and GPT-4o, it's evident from Figure 7 that both models exhibit similar NRA values, indicating comparable understanding and response accuracy to non-adversarial prompts under normal conditions. However, Figure 6 highlights a significant disparity in their robustness, with GPT-4-turbo outperforming GPT-4o in adversarial situations.

The following discussion will focus on the RRA values, as shown in Figure 8. These values measure the accuracy of responses to adversarial prompts using the T2P module and few-shot attack strategies. Despite a general decline in accuracy under adversarial attacks for both models, GPT-4-turbo consistently maintains higher accuracy than GPT-4o across different datasets, highlighting its superior robustness..

\begin{figure}[h]
    \centering
    \includegraphics[width=1 \linewidth]{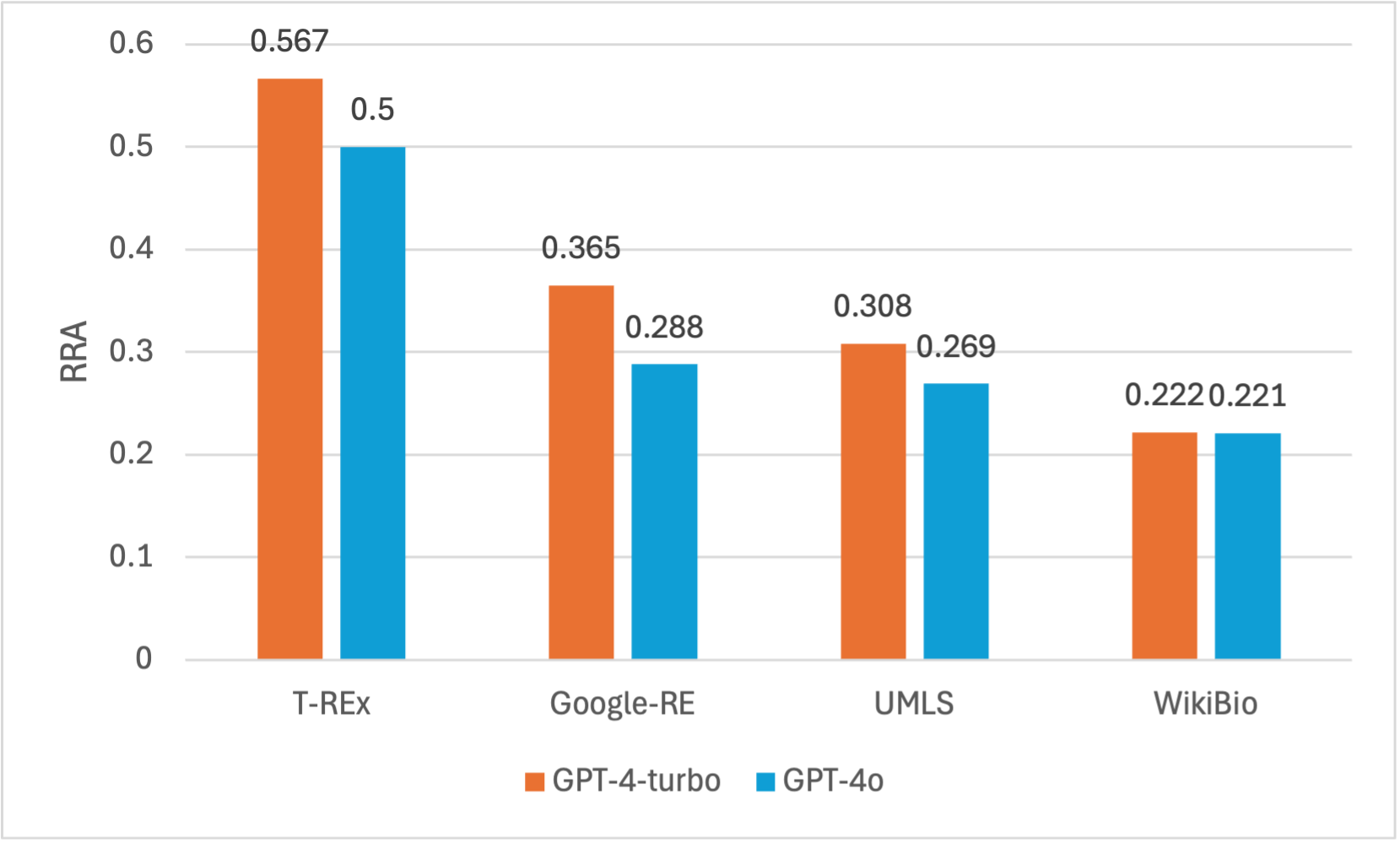}
    \caption{RRA values of GPT-4-turbo and GPT-4o on various knowledge graphs}
    \label{fig:enter-label}
\end{figure}
This comparison undermines the hypothesis that prompts answered correctly under adversarial conditions would similarly be answered correctly under normal scenarios. For instance, the RRA of GPT-4-turbo in the T-REx dataset is 0.567, slightly lower than its NRA of 0.606. This discrepancy, alongside an actual ASR of 0.048, lower than the 0.064 expected under the hypothesis, suggests that some prompts may yield different responses under normal and adversarial conditions, highlighting an inherent randomness in large language model responses. Nonetheless, these metrics effectively demonstrate the robustness of large language models and the relative strengths between different models.
\begin{figure}[h]
    \centering
    \includegraphics[width=1\linewidth]{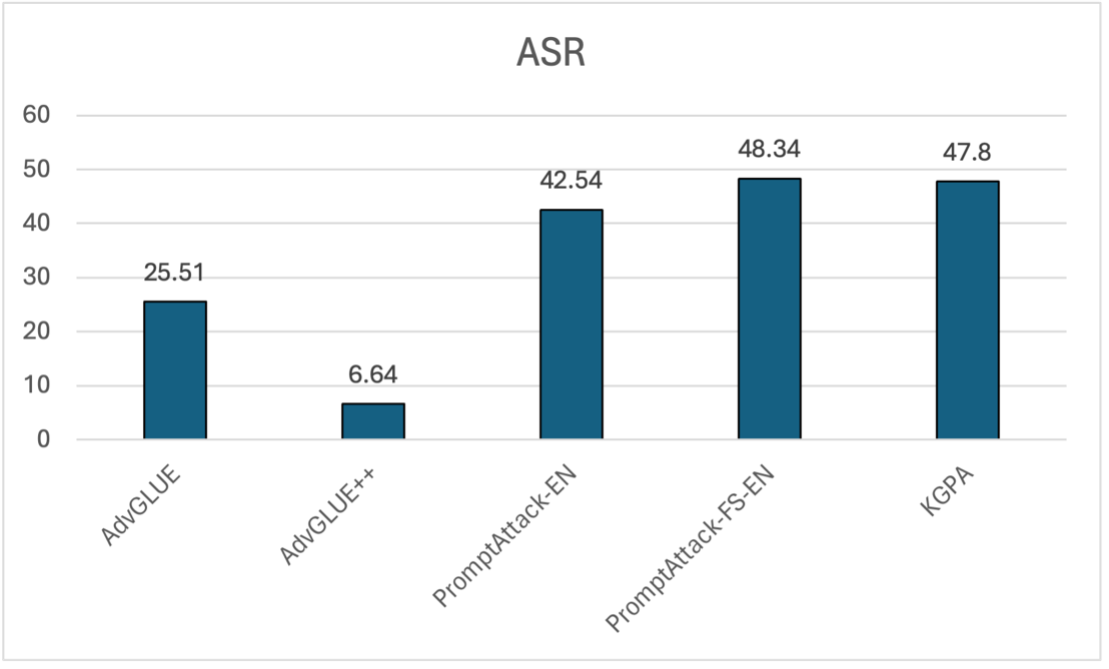}
    \caption{ASR values obtained by each framework from robustness attacks on GPT-3.5}
    \label{fig:enter-label}
\end{figure}




\noindent\textbf{Comparison Analysis.} This section evaluates the KGPA framework against other robustness frameworks like AdvGLUE, AdvGLUE++, PromptAttack-EN, and PromptAttack-FS-EN, focusing on three key aspects: adversarial success rates (ASR), reliability, and usage costs.According to Figure 9, ASR of KGPA for GPT-3.5 is slightly lower than that of PromptAttack-FS-EN, but higher than PromptAttack-EN and significantly higher than both versions of AdvGLUE. This demonstrates KGPA’s effectiveness in generating challenging adversarial samples and in robustness evaluation, offering a broad spectrum for assessing model robustness.

\noindent\textbf{Accuracy.} Conclusions for KGPA align with the parameter sizes and actual performance of models when evaluating GPT-3.5-turbo, GPT-4-turbo, and GPT-4o, i.e., GPT-4-turbo > GPT-4o > GPT-3.5-turbo. In contrast, other frameworks incorrectly concluded that Llama2-13B is less robust than Llama2-7B, despite Llama2-13B having nearly twice the parameters.

\noindent\textbf{Usage Cost.} AdvGLUE and similar frameworks incur high usage costs due to their reliance on manually annotated benchmark datasets, which may not accurately reflect robustness in specialized domains. Conversely, KGPA reduces costs by using automatically constructed knowledge graph datasets, providing broad applicability across various domains.

In summary, compared to other existing robustness evaluation frameworks, KGPA not only has lower usage costs but also demonstrates excellent performance in evaluating large language model robustness.

\subsection{Experimental Analysis of KGPA Modules} 

This analysis synthesizes outcomes from three KGPA components to assess their combined effect on model robustness. It explores prompt strategies of T2P, evaluates the few-shot efficacy of attack, and examines PRE module thresholds, underscoring their impact on accuracy and adversarial responses. Module of PRE threshold settings, highlighting their impact on adversarial attack and robustness evaluation.

\noindent{}\textbf{T2P Strategies: Template vs. LLM-Based.} We implemented a few-shot attack strategy with the module of PRE threshold set to 0.92, yielding ASR values for both template and LLM-based strategies in the T2P module as shown in Table 1. Analysis of Table 1 indicates that for the same knowledge graph dataset, ASR values for three LLMs rank as GPT-3.5-turbo > GPT-4o > GPT-4-turbo with unchanged generation strategies in the T2P module. Furthermore, T2P modules with LLM-based strategies outperformed template-based ones, suggesting that LLM-based strategies are more effective for robust LLM evaluation and adversarial attacks.

\begin{table*}[h]
\centering
\label{tab:full-width-table}
\begin{tabular}{lccccc}
\toprule
\textbf{Model} & \textbf{T2P} & \textbf{T-REx} & \textbf{Google-RE} & \textbf{UMLS} & \textbf{WikiBio} \\
\midrule
\multirow{2}{*}{\textbf{GPT-3.5-turbo}} & \textbf{LLM-Based} & \textcolor{red}{0.448} & \textcolor{red}{0.737} & \textcolor{red}{0.455} & \textcolor{red}{0.600} \\
 & \textbf{Template-Based} & 0.419 & 0.391 & 0.231 & 0.444 \\
\midrule
\multirow{2}{*}{\textbf{GPT-4-turbo}} & \textbf{LLM-Based} & \textcolor{red}{0.063} & 0.095 & \textcolor{red}{0.347} & \textcolor{red}{0.267} \\
 & \textbf{Template-Based} & \textcolor{red}{0.063} & \textcolor{red}{0.159} & 0.162 & 0.206 \\
\midrule
\multirow{2}{*}{\textbf{GPT-4o}} & \textbf{LLM-Based} & 0.133 & \textcolor{red}{0.492} & \textcolor{red}{0.417} & \textcolor{red}{0.361} \\
 & \textbf{Template-Based} & \textcolor{red}{0.250} & 0.364 & 0.341 & 0.333 \\
\bottomrule
\end{tabular}
\caption{ASR: Comparison of LLM-based and template-based strategies across knowledge graphs}
\end{table*}

\noindent{}\textbf{Few-Shot Attack Strategy: Adversarial Impact.} We categorize knowledge graph datasets into general (T-REx, Google-RE) and specialized domains (UMLS for medicine, WikiBio for biology). By calculating average ASR values, we assess the robustness of each large language model across these datasets, summarized in Figures 10 and 11. "FSA" denotes inclusion of few-sample attack strategies, while "NO-FSA" indicates their absence. The graphs show that the KGPA framework achieves higher ASR values with FSA than without. However, GPT-4o scores higher on specialized domain datasets without FSA, suggesting better suitability for general domains.

\begin{figure}[h]
        \centering
        \includegraphics[width=1\linewidth]{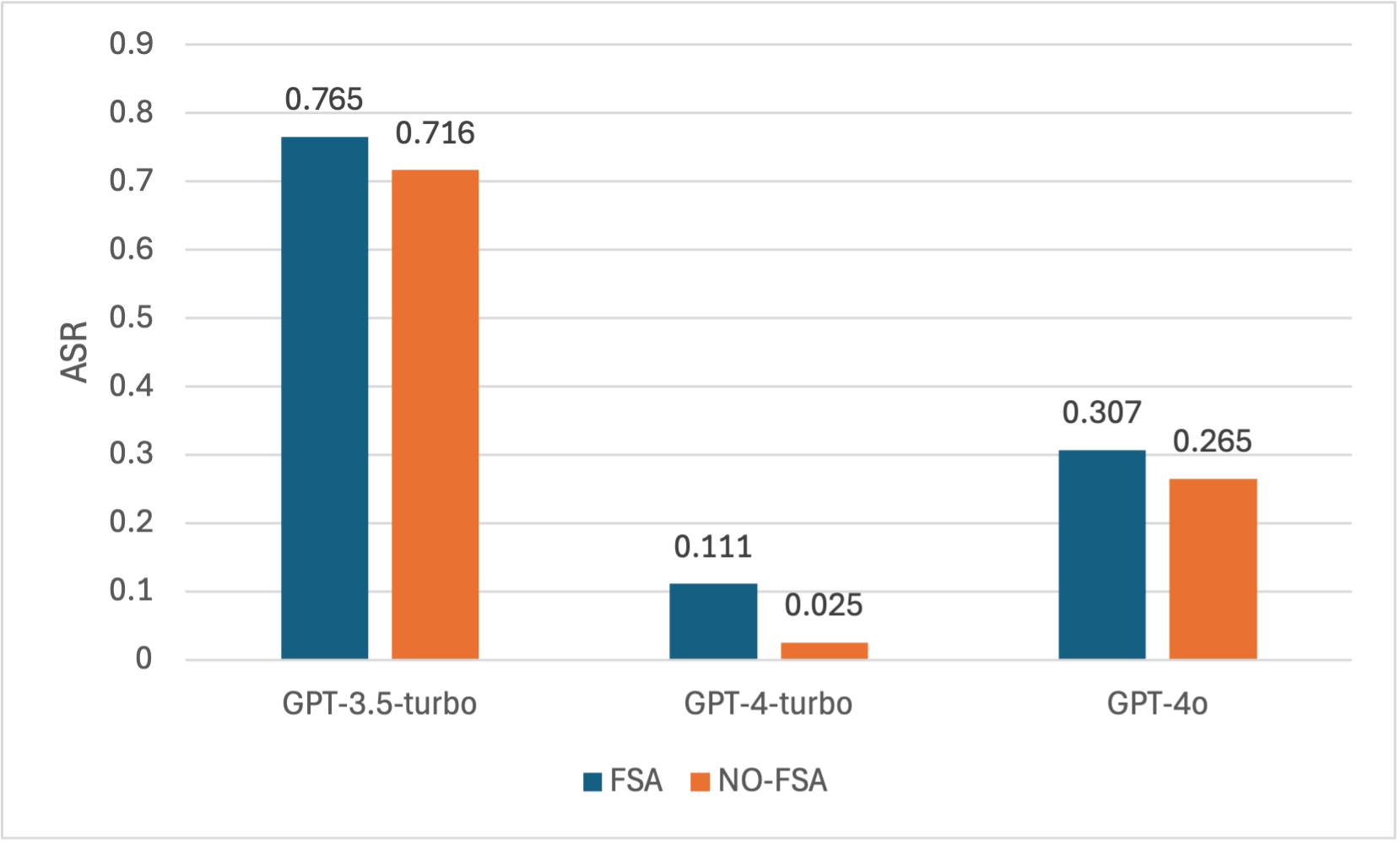}
        \caption{ASR comparison of LLMs on general domains with and without few-sample attack strategies}
        \label{fig:enter-label}
\end{figure}

\begin{figure}[h]
        \centering
        \includegraphics[width=1\linewidth]{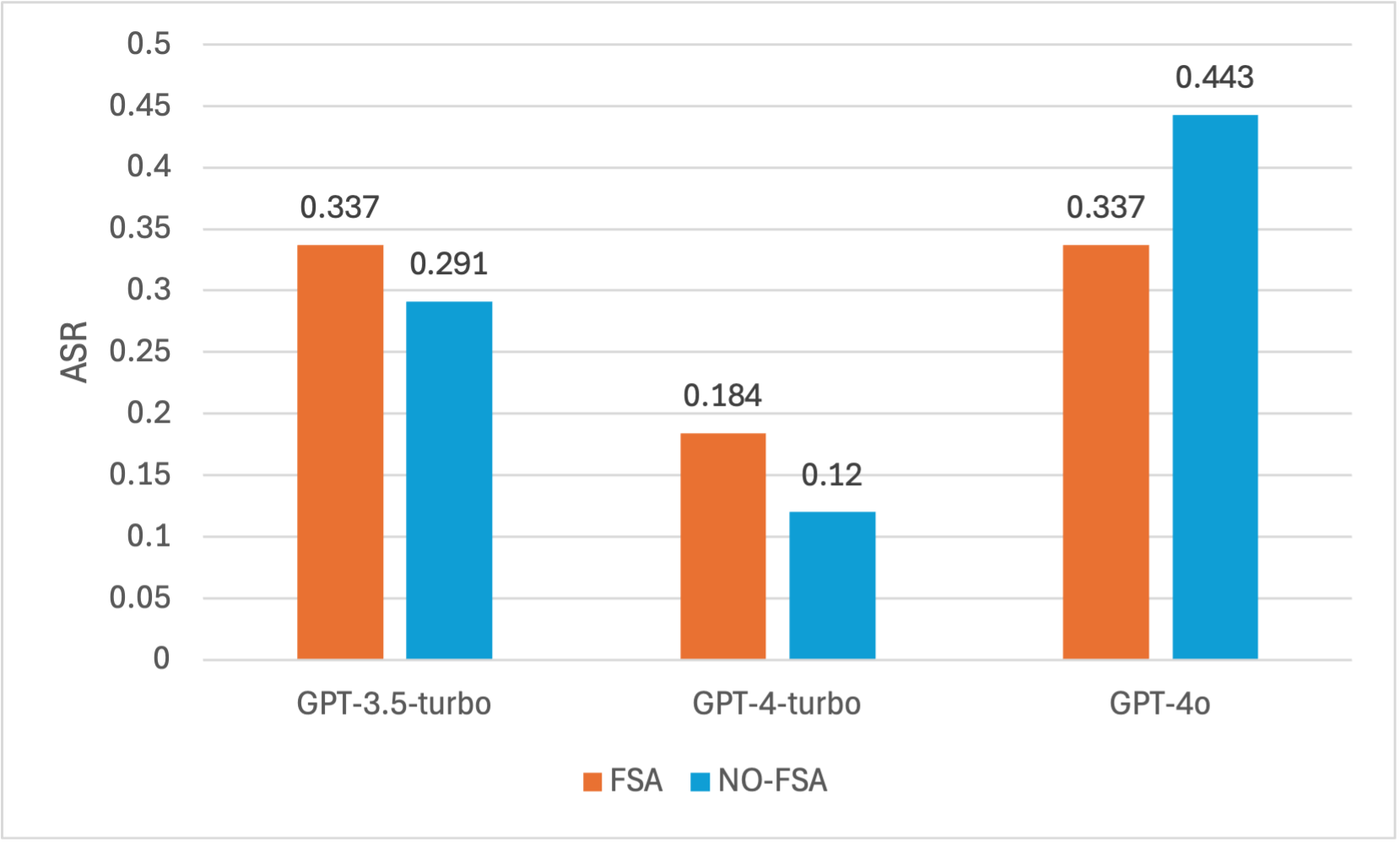}
        \caption{ASR comparison of LLMs on specialized domains with/without few-sample attack strategies}
        \label{fig:enter-label}
\end{figure}

\noindent{}\textbf{PRE Threshold: Effects on Robustness Evaluation.} The KGPA framework uses the PRE module to filter out low-quality or altered prompts from the APGP module, ensuring high-quality adversarial samples. Figure 12 shows that with tau\_llm below 0.9, ASR differences between GPT-3.5-turbo and GPT-4.0-turbo are minimal. Raising tau\_llm above 0.9 widens the ASR gap, stabilizing ASR of GPT-3.5-turbo and decreasing GPT-4-turbo. Based on Figure 13, this occurs because removing low-quality prompts makes adversarial prompts of  GPT-4-turbo closer to the original. Since the LLM used by the PRE module to calculate the LLMScore is the same LLM being evaluated for robustness, it reflects the corresponding LLM’s robustness. Thus, tau\_llm was set at 0.92 for most experiments.

\begin{figure}[h]
        \centering
        \includegraphics[width=1\linewidth]{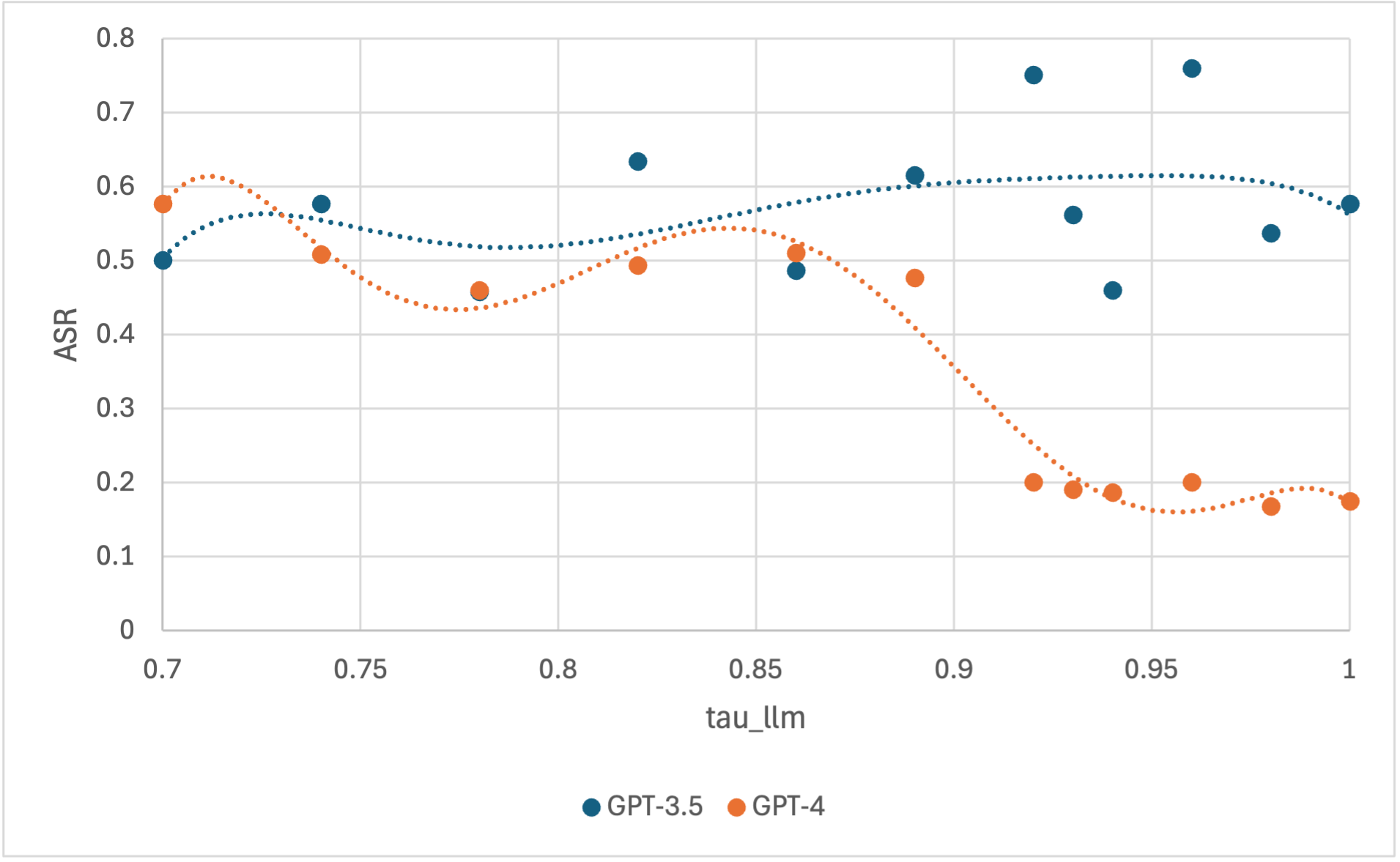}
        \caption{ASR values at different tau\_llm thresholds (Dashed line: Fitted curve)}
        \label{fig:enter-label}
\end{figure}

\begin{figure}[h]
        \centering
        \includegraphics[width=1\linewidth]{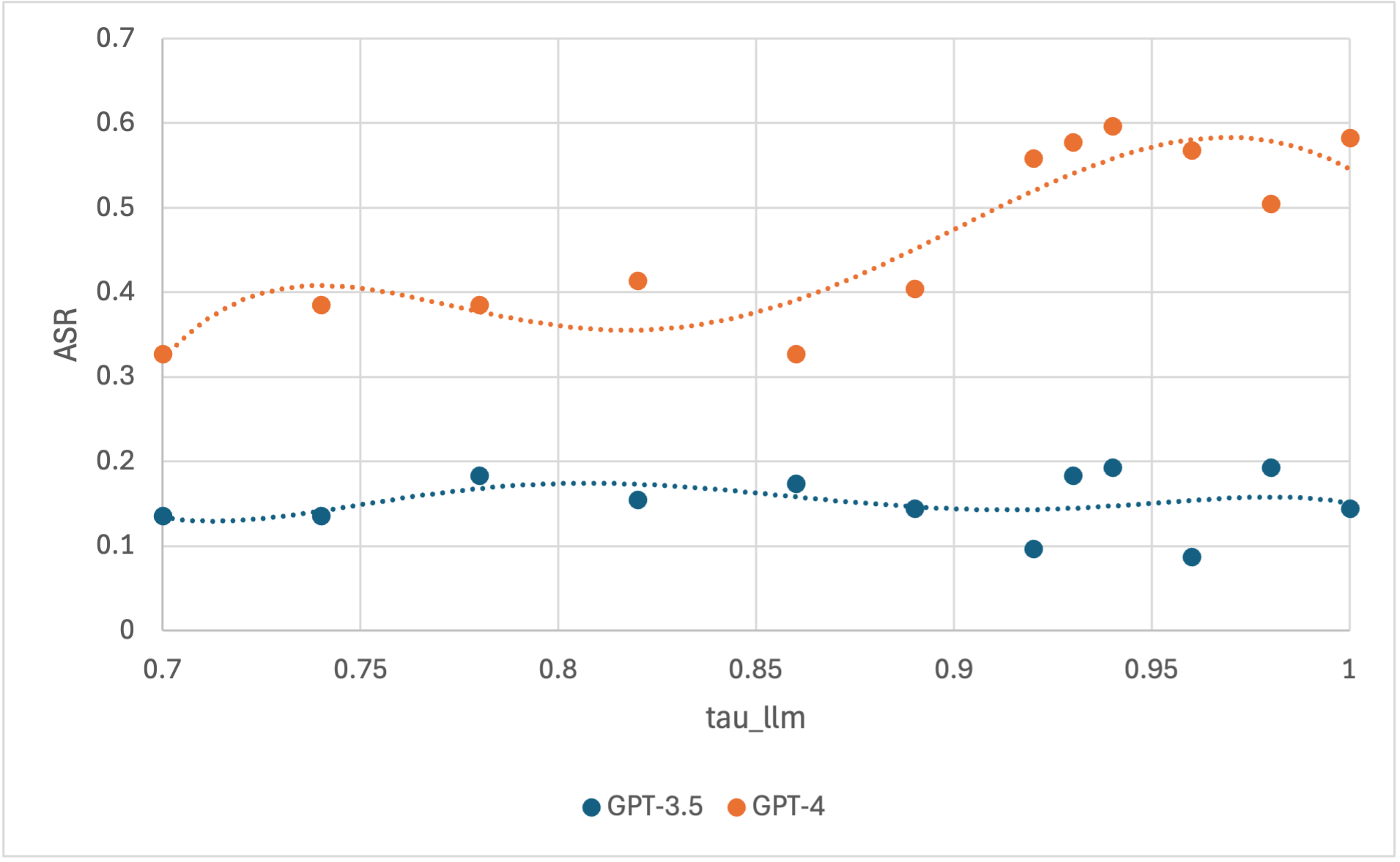}
        \caption{RRA values at different tau\_llm thresholds (Dashed line: Fitted curve)}
        \label{fig:enter-label}
\end{figure}

\section{Conclusion}

We propose a knowledge graph-based systematic framework to evaluate the robustness of large language models (LLMs) in adversarial attack environments across different domains. The experiments assessed the robustness of several models in the ChatGPT family and analyzed the factors affecting the Adversarial Success Rate (ASR) within the framework, studying the differences in LLM robustness across various domains. Our research contributes to the study of LLM robustness evaluation and adversarial attacks.

\section*{Limitations}

The limitations of our work includes:

\begin{itemize}

\item Types of problems for evaluating LLM robustness. In the KGPA framework, we require the LLM to perform classification tasks to assess the robustness of large language models; in future research, we plan to enrich the types of problems by including types such as short answer questions and true/false questions, to conduct a more comprehensive evaluation of the LLM of robustness.

\item Details of the PRE Module. In the KGPA framework, the PRE module is designed to filter qualified adversarial prompts based on the large language model’s self-assessment score, LLMScore. As the current scoring method is somewhat rudimentary, we intend to refine the scoring criteria and evaluate appropriate threshold settings to better tailor the LLMScore to the needs of the PRE module’s filtering tasks.

\end{itemize}

\bibliography{custom}

\appendix

\clearpage 
\section{Experimentation Details}

\subsection{Dataset}

In this experiment, we divided the knowledge graph dataset into two categories based on the scope of knowledge represented by the knowledge graphs, including the general domain knowledge graph datasets and the specialized domain knowledge graph datasets. The general domain knowledge graph datasets include T-REx and Google-RE, while the specialized domain knowledge graph datasets include UMLS and WikiBio.

\begin{itemize}

\item T-REx \cite{elsahar-etal-2018-rex}: Originating from Wikipedia, this is a general domain knowledge graph that records a large number of predicates and entities belonging to various fields.
\item Google-RE \cite{petroni-etal-2019-language}: The dataset information, sourced from Wikipedia, primarily records the birth and death locations and birthdates of various individuals. Although this information may make the dataset appear to be a specialized domain knowledge graph, we consider it a general domain knowledge graph due to the widely known nature of the information included.
\item UMLS \cite{Bodenreider2004TheUM}: This is a specialized knowledge graph in the medical field, constructed by experts in the domain, and it contains information about various medical concepts and their relationships.
\item WikiBio \cite{sung2021can}: This dataset is constructed by extracting biology-related instances from Wikidata—one of the largest knowledge graphs in the world—and is a specialized knowledge graph in the field of biology.

\end{itemize}

\subsection{Implementations}

\textbf{Large Language Model.} We have utilized several models from the ChatGPT family, including GPT-3.5-turbo, GPT-4-turbo, and GPT-4o. The large language models were accessed via paid APIs to complete relevant robustness evaluation tasks.

\noindent{}\textbf{Prompt Generation and Response Processing.} We set the ratio of the three labels "true," "entity\_error," and "predicate\_error" for the generated prompts to 1:1:1. To extract the classification results from responses of the LLM for the classification task, we employed string matching. If a response matches one of the aforementioned three labels and the label is the correct one, classification of the LLM is deemed correct; otherwise, it is considered incorrect.

\section{Original Experimental Results}

In Tables 2 to 10, we present partial results of the experiments. The labels of the tables include the name of the data (ASR, NRA, or RRA), the Original Prompt generation strategy used in the T2P module (template-based and LLM-based), and whether a few-shot attack strategy is employed (FSA: Yes; NO-FSA: No). The threshold tau\_llm for the PRE module is set at 0.92. In the GPT column, 3.5, 4, and 4o respectively represent GPT-3.5-turbo, GPT-4-turbo, and GPT-4o.

\begin{table}[H]
\centering
\label{tab:full-width-table}
\begin{tabular}{lcccc}
\toprule
\textbf{GPT} & \textbf{T2P} & \textbf{T-REx} & \textbf{Google-RE} & \textbf{UMLS} \\
\midrule
\textbf{3.5} & 0.448 & 0.737 & 0.455 & 0.600 \\
\textbf{4} & 0.063 & 0.095 & 0.347 & 0.267 \\
\textbf{4o} & 0.133 & 0.492 & 0.417 & 0.361 \\
\bottomrule
\end{tabular}
\caption{ASR: T2P: LLM-based \&\& FSA}
\end{table}

\begin{table}[H]
\centering
\label{tab:full-width-table}
\begin{tabular}{lcccc}
\toprule
\textbf{GPT} & \textbf{T2P} & \textbf{T-REx} & \textbf{Google-RE} & \textbf{UMLS} \\
\midrule
\textbf{3.5} & 0.779 & 0.751 & 0.231 & 0.444 \\
\textbf{4} & 0.062 & 0.159 & 0.162 & 0.206 \\
\textbf{4o} & 0.250 & 0.364 & 0.341 & 0.333 \\
\bottomrule
\end{tabular}
\caption{ASR: T2P: Template-based \&\& FSA}
\end{table}

\begin{table}[H]
\centering
\label{tab:full-width-table}
\begin{tabular}{lcccc}
\toprule
\textbf{GPT} & \textbf{T2P} & \textbf{T-REx} & \textbf{Google-RE} & \textbf{UMLS} \\
\midrule
\textbf{3.5} & 0.828 & 0.604 & 0.524 & 0.059 \\
\textbf{4} & 0.018 & 0.032 & 0.140 & 0.100 \\
\textbf{4o} & 0.143 & 0.387 & 0.400 & 0.486 \\
\bottomrule
\end{tabular}
\caption{ASR: T2P: Template-based \&\& NO-FSA}
\end{table}

\begin{table}[H]
\centering
\begin{tabular}{lcccc}
\toprule
\textbf{GPT} & \textbf{T2P} & \textbf{T-REx} & \textbf{Google-RE} & \textbf{UMLS} \\
\midrule
\textbf{3.5} & 0.279 & 0.183 & 0.212 & 0.192 \\
\textbf{4} & 0.606 & 0.604 & 0.471 & 0.288 \\
\textbf{4o} & 0.577 & 0.567 & 0.462 & 0.346 \\
\bottomrule
\end{tabular}
\caption{NRA: T2P: LLM-based \&\& FSA}
\end{table}

\begin{table}[H]
\centering
\label{tab:full-width-table}
\begin{tabular}{lcccc}
\toprule
\textbf{GPT} & \textbf{T2P} & \textbf{T-REx} & \textbf{Google-RE} & \textbf{UMLS} \\
\midrule
\textbf{3.5} & 0.298 & 0.442 & 0.250 & 0.087 \\
\textbf{4} & 0.615 & 0.606 & 0.356 & 0.327 \\
\textbf{4o} & 0.615 & 0.529 & 0.423 & 0.288 \\
\bottomrule
\end{tabular}
\caption{NRA: T2P: Template-based \&\& FSA}
\end{table}

\begin{table}[H]
\centering
\label{tab:full-width-table}
\begin{tabular}{lcccc}
\toprule
\textbf{GPT} & \textbf{T2P} & \textbf{T-REx} & \textbf{Google-RE} & \textbf{UMLS} \\
\midrule
\textbf{3.5} & 0.278 & 0.461 & 0.202 & 0.163 \\
\textbf{4} & 0.548 & 0.596 & 0.481 & 0.288 \\
\textbf{4o} & 0.606 & 0.596 & 0.481 & 0.337 \\
\bottomrule
\end{tabular}
\caption{NRA: T2P: Template-based \&\& NO-FSA}
\end{table}

\begin{table}[H]
\centering
\label{tab:full-width-table}
\begin{tabular}{lcccc}
\toprule
\textbf{GPT} & \textbf{T2P} & \textbf{T-REx} & \textbf{Google-RE} & \textbf{UMLS} \\
\midrule
\textbf{3.5} & 0.154 & 0.048 & 0.115 & 0.077 \\
\textbf{4} & 0.567 & 0.365 & 0.308 & 0.222 \\
\textbf{4o} & 0.500 & 0.288 & 0.269 & 0.222 \\
\bottomrule
\end{tabular}
\caption{RRA: T2P: LLM-based \&\& FSA}
\end{table}

\begin{table}[H]
\centering
\label{tab:full-width-table}
\begin{tabular}{lcccc}
\toprule
\textbf{GPT} & \textbf{T2P} & \textbf{T-REx} & \textbf{Google-RE} & \textbf{UMLS} \\
\midrule
\textbf{3.5} & 0.173 & 0.269 & 0.192 & 0.048 \\
\textbf{4} & 0.577 & 0.510 & 0.298 & 0.260 \\
\textbf{4o} & 0.462 & 0.337 & 0.279 & 0.192 \\
\bottomrule
\end{tabular}
\caption{RRA: T2P: Template-based \&\& FSA}
\end{table}

\begin{table}[H]
\centering
\label{tab:full-width-table}
\begin{tabular}{lcccc}
\toprule
\textbf{GPT} & \textbf{T2P} & \textbf{T-REx} & \textbf{Google-RE} & \textbf{UMLS} \\
\midrule
\textbf{3.5} & 0.048 & 0.183 & 0.096 & 0.154 \\
\textbf{4} & 0.538 & 0.577 & 0.413 & 0.260 \\
\textbf{4o} & 0.519 & 0.365 & 0.288 & 0.173 \\
\bottomrule
\end{tabular}
\caption{RRA: T2P: Template-based \&\& NO-FSA}
\end{table}

\section{Prompt Templates}
In this section, we introduce the prompt templates used in the KGPA framework, including those in the T2P module, KGB-FSA module, PRE module, APGP module, and the Robustness Evaluation part, and explain their meanings.

\subsection{T2P\_prompt}

The "T2P\_prompt" is a prompt template in the T2P module as shown in Figure 2, incorporating triplet components to guide large language model outputs, distinct from template-based strategies. Highlighted in red in Table 11, specific template parts require replacement with actual triplet data. This method enables LLM-based strategy in T2P module to convert knowledge graph triplets into syntactically correct, natural-sounding sentences by leveraging the large model.

\begin{table*}[ht]
\centering
\label{tab:full-width-table}
\begin{tabular}{l}
\toprule
\textbf{T2P\_prompt} \\
\midrule
Here is a triplet (subject, predicate, object) extracted from knowledge graph: \\
The subject: \textcolor{red}{Subject}; The predicate: \textcolor{red}{Predicate}; The object: \textcolor{red}{Object}; \\
Now create a statement describing this triplet. Tips: Do not care about whether this triplet is true, and do \\
not change the meaning of the predicate. \\
Just give the statement. Statement: \\
\bottomrule
\end{tabular}
\caption{T2P\_prompt: Generating original prompts}
\end{table*}

\subsection{KGB-FSA\_prompt}
The "KGB-FSA\_prompt" (referenced in Table 12) instructs the model to modify a sentence to retain semantic consistency with the original while aiming for different classification outcomes. The template highlights replacement parts in red: "\textcolor{red}{Sentence}" for the original sentence, "\textcolor{red}{Ground Truth Label}" for the actual label, and "\textcolor{red}{Other Labels}" for alternative labels. Success is achieved if the modification preserves the original meaning but potentially causes classification discrepancies.

\begin{table*}[ht]
\label{tab:full-width-table}
\centering
\begin{tabular}{l}
\toprule
\textbf{KGB-FSA\_prompt} \\
\midrule
The original sentence "\textcolor{red}{Sentence}" is classified as \textcolor{red}{Ground Truth Label}. \\
Your task is to generate a new sentence which must satisfy the following  conditions: \\
1. Keeping the semantic meaning of the new sentence unchanged; \\
2. The new sentence should be classified as \textcolor{red}{Other Labels} \\
You can finish the task by modifying sentence using the following guidance: Paraphrase the sentence ; \\
Only output the new sentence without anything else. \textcolor{red}{Sentence} -> \\
\bottomrule
\end{tabular}
\caption{KGB-FSA\_prompt: Generating adversarial prompt examples}
\end{table*}

\subsection{PRE\_prompt}

Combine the generated sentence from a large language model (Generated Sentence) with the original sentence (Original Sentence) using the PRE\_prompt template (as shown in Table 13), and input them together into the large language model. The model evaluates the semantic similarity and quality of the generated sentence to produce a score ranging from -1 to 1, called the LLMScore. A higher LLMScore indicates greater semantic similarity and higher quality of the generated sentence. The parts in red need to be replaced with the original and generated sentences.

\begin{table*}[ht]
\label{tab:full-width-table}
\centering
\begin{tabular}{l}
\toprule
\textbf{PRE\_prompt} \\
\midrule
Task Description: \\
you are required to assess the quality of a piece of generated text, as well as its semantic similarity and \\
overall quality in relation to a provided reference text. \\
Input Data:
Reference Text: \textcolor{red}{Original Sentence}; Candidate Text: \textcolor{red}{Generated Sentence} \\
Scoring Guidelines: Please rate the candidate text according to the following criteria, with scores \\
ranging from -1 to 1, where 1 represents a perfect match, and -1 represents a complete mismatch. The \\
score can be a decimal \\
Just give the score, and the Score needs to be a three-digit decimal representation, score: \\
\bottomrule
\end{tabular}
\caption{PRE\_prompt: Generating LLMScore}
\end{table*}

\subsection{APGP\_prompt}

The APGP\_prompt (as shown in Table 14) is similar to the KGB-FSA\_prompt, requiring the LLM to generate adversarial prompts. The distinction lies in the APGP\_prompt offering an optional few-shot Attack Strategy, which allows for the inclusion of example prompts within APGP\_prompt to guide the large model in generating results.

\begin{table*}[ht]
\label{tab:full-width-table}
\centering
\begin{tabular}{l}
\toprule
\textbf{APGP\_prompt} \\
\midrule
The original sentence "\textcolor{red}{Sentence}" is classified as \textcolor{red}{Ground Truth Label}. \\
Your task is to generate a new sentence which must satisfy the following  conditions: \\
1. Keeping the semantic meaning of the new sentence unchanged; \\
2. The new sentence should be classified as \textcolor{red}{Other Labels} \\
You can finish the task by modifying sentence using the following guidance: Paraphrase the sentence; \\
(Here is an example that fit the guidance: Few-shot Example Original Sentence ->  Few-shot Example \\
Generated Sentence) \\
Only output the new sentence without anything else. \\
\textcolor{red}{Sentence} -> \\
\bottomrule
\end{tabular}
\caption{APGP\_prompt: Generating adversarial prompts}
\end{table*}

\subsection{Prompt Template Used in Robustness Evaluation Part}

Using the prompts shown in Table 15, require the LLM to perform the classification task as specified in the prompt, and provide the classification results.

\begin{table*}[ht]
\label{tab:full-width-table}
\centering
\begin{tabular}{l}
\toprule
\textbf{Prompt Template Used in Robustness Evaluation Part} \\
\midrule
Appraise the category of a sentence derived from a triplet (subject, predicate, object) and decide if it is \\
'true', 'entity\_error' or 'predicate\_error': \\
\textcolor{red}{Sentence} \\
\bottomrule
\end{tabular}
\caption{Prompt template used in robustness evaluation part}
\end{table*}

\end{document}